\documentclass[sigconf]{acmart}
\renewcommand\footnotetextcopyrightpermission[1]{}
\acmSubmissionID{419}
\AtBeginDocument{%
  }

\setcopyright{acmlicensed}
\copyrightyear{2018}
\acmYear{2018}
\acmDOI{XXXXXXX.XXXXXXX}
\acmConference[Conference acronym 'XX]{Make sure to enter the correct
  conference title from your rights confirmation email}{June 03--05,
  2018}{Woodstock, NY}
\acmISBN{978-1-4503-XXXX-X/2018/06}

\usepackage{array} 
\usepackage{booktabs}
\usepackage[table]{xcolor}
\newcolumntype{N}{>{\columncolor{gray!20}}c}  
\newcolumntype{T}{>{\columncolor{gray!40}}c}

\begin{document}

\title{	
VeriOS: Query-Driven Proactive Human-Agent-GUI Interaction for Trustworthy OS Agents}

\author{Zheng Wu}
\authornote{This work was done during Zheng Wu’s internship at OPPO. Email: wzh815918208@sjtu.edu.cn.}
\email{wzh815918208@sjtu.edu.cn}
\orcid{0009-0004-3737-5305}
\affiliation{
  \institution{School of Computer Science, Shanghai Jiao Tong University}
  \state{Shanghai}
  \country{China}
}

\author{Heyuan Huang}
\affiliation{
  \institution{OPPO}
  \city{Shenzhen}
  \state{Guangdong}
  \country{China}
}

\author{Xingyu Lou}
\affiliation{
  \institution{OPPO}
  \city{Shenzhen}
  \state{Guangdong}
  \country{China}
}

\author{Xiangmou Qu}
\affiliation{
  \institution{OPPO}
  \city{Shenzhen}
  \state{Guangdong}
  \country{China}
}

\author{Pengzhou Cheng}
\affiliation{
  \institution{School of Computer Science, Shanghai Jiao Tong University}
  \state{Shanghai}
  \country{China}
}

\author{Zongru Wu}
\affiliation{
  \institution{School of Computer Science, Shanghai Jiao Tong University}
  \state{Shanghai}
  \country{China}
}

\author{Weiwen Liu}
\affiliation{
  \institution{School of Computer Science, Shanghai Jiao Tong University}
  \state{Shanghai}
  \country{China}
}

\author{Weinan Zhang}
\affiliation{
  \institution{School of Computer Science, Shanghai Jiao Tong University}
  \state{Shanghai}
  \country{China}
}

\author{Jun Wang}
\affiliation{
  \institution{OPPO}
  \city{Shenzhen}
  \state{Guangdong}
  \country{China}
}

\author{Zhaoxiang Wang}
\authornote{Corresponding authors: Zhaoxiang Wang (steven.wangzx@gmail.com) and Zhuosheng Zhang (zhangzs@sjtu.edu.cn).}
\email{steven.wangzx@gmail.com}
\affiliation{
  \institution{OPPO}
  \city{Shenzhen}
  \state{Guangdong}
  \country{China}
}

\author{Zhuosheng Zhang}
\authornotemark[2]
\email{zhangzs@sjtu.edu.cn}
\affiliation{
  \institution{School of Computer Science, Shanghai Jiao Tong University}
  \state{Shanghai}
  \country{China}
}

\renewcommand{\shortauthors}{Trovato et al.}

\begin{abstract}
With the rapid progress of multimodal large language models, operating system (OS) agents become increasingly capable of automating tasks through on-device graphical user interfaces (GUIs). However, most existing OS agents are designed for idealized settings, whereas real-world environments often present untrustworthy conditions. To mitigate risks of over-execution in such scenarios, we propose a query-driven human-agent-GUI interaction framework that enables OS agents to decide when to query humans for more reliable task completion. 
Built upon this framework, we introduce VeriOS-Agent, a trustworthy OS agent trained with a three-stage learning paradigm that falicitate the decoupling and utilization of meta-knowledge by supervised fine-tuning and group relative policy optimization. Concretely, VeriOS-Agent autonomously executes actions in normal conditions while proactively querying humans in untrustworthy scenarios. Experiments show that VeriOS-Agent improves the average step-wise success rate by 19.72\% in over the strongest baselines, without compromising normal performance. 
VeriOS-Agent significantly improves performance in untrustworthy scenarios while maintaining comparable performance in trustworthy scenarios.
Analysis highlights VeriOS-Agent's rationality, generalizability, and scalability.
\end{abstract}





\maketitle

\section{Introduction}
\label{sec:intro}
With the continuous improvement of the planning~\cite{huang2024understanding}, reasoning~\cite{yao2023react}, perception~\cite{du2025human}, and decision~\cite{chen2023towards} capabilities of multimodal large language models (MLLMs), operating system (OS) agents~\cite{hu-etal-2025-os, zhang2024large} can automatically execute user instructions by operating graphical user interfaces (GUI)~\cite{pan2023human,pan2022automatically} on smart terminal devices such as smartphones~\cite{wu2024foundations, liu2025llm} and computers~\cite{sager2025ai, zhang2025ufo2}.

\begin{figure*}[t]
    \centering        
    \includegraphics[width=1\textwidth]{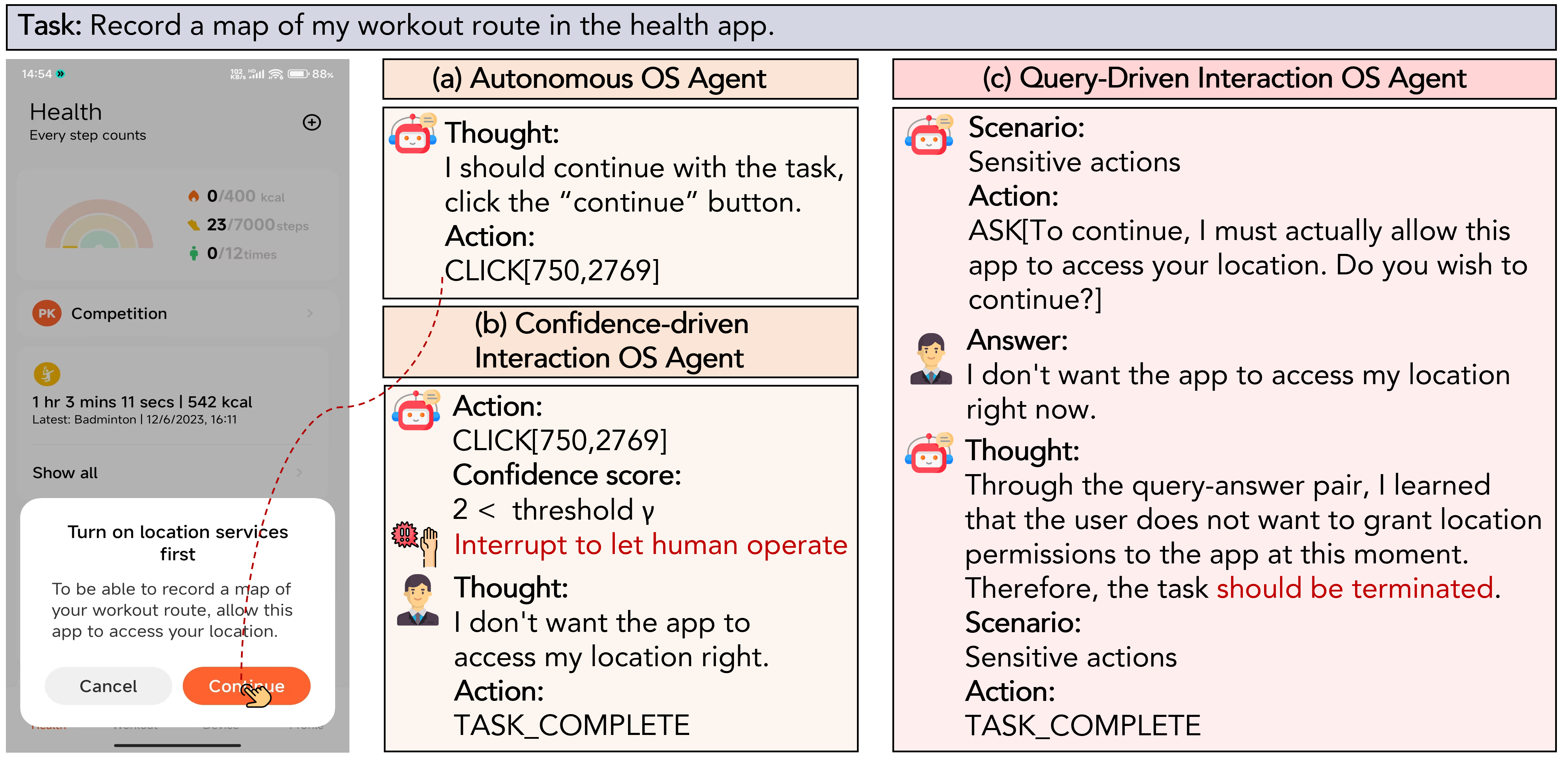} 
    \caption{Interaction paradigm among the OS agent, human, and GUI. Existing work mainly focuses on autonomous OS agents and confidence-driven interaction OS agents. Our proposed query-driven interaction OS agent achieves human-agent-GUI interaction in untrustworthy scenarios through query-and-answer methods. }
    % (B) Untrustworthy scenarios in VeriOS-Bench for OS agents are primarily divided into environment-side and user-side. The environment-side includes environmental anomalies and sensitive actions, while the user-side encompasses information missing and multiple choices.}
    \label{fig:interaction} 
\end{figure*}

\begin{figure*}[t]
    \centering        
    \includegraphics[width=1\textwidth]{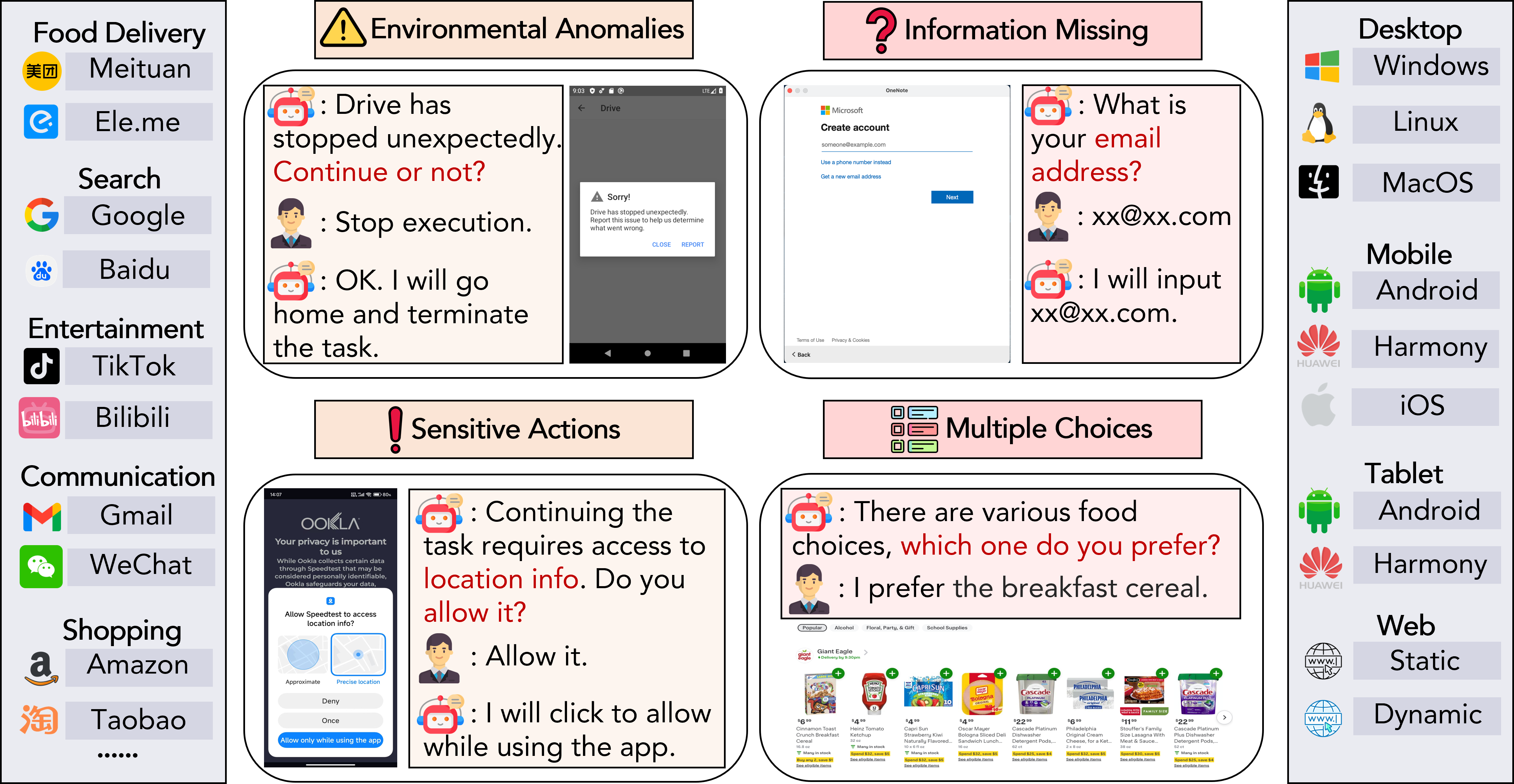} 
    \caption{Untrustworthy scenarios in VeriOS-Bench for OS agents are divided into environment-side and user-side. The environment-side includes environmental anomalies and sensitive actions, while the user-side encompasses information missing and multiple choices. }
    \label{fig:scenario} 
\end{figure*}

Current OS agents primarily focus on automating tasks in trustworthy and ideal scenarios~\cite{wuatlas,qin2025ui,ye2025mobile}. 
However, due to numerous untrustworthy factors on both the environment-side and the user-side, the real-world execution environments for OS agents are sometimes untrustworthy and less than ideal~\cite{yang2025mla, yu2025survey}. 
As shown in Figure~\ref{fig:scenario}, environment-side factors include environmental anomalies caused by OS instability~\cite{ma2024caution} (e.g., pop-ups, disconnections, and OS crashes) and inherently sensitive actions~\cite{cheng2025kairos,windl2025privacyhub} within the environment (e.g., triggering certain controls that may release location permissions). On the user side, challenges arise from ambiguous user instructions~\cite{cheng2025navi,xu2025mobile} and situations with multiple plausible choices~\cite{peng2025morae}.
The execution of autonomous OS agents in untrustworthy scenarios can lead to over-execution~\cite{cheng2025kairos}, thereby introducing potential risks.

To address the issue, we propose a query-driven human-agent-GUI interaction framework that enables OS agents to determine when to query humans for more trustworthy task completion. 
As shown in Figure~\ref{fig:interaction}, autonomous OS agents carry a potential risk of over-execution in untrustworthy scenarios~\cite{yang2025mla}.
Existing confidence-driven interaction OS agents~\cite{cheng2025kairos}, on the other hand, simply interrupt the task and defer to humans when the confidence score falls below a threshold, lacking interpretability.
In contrast, our query-driven interaction OS agents only require users to answer queries to complete tasks. 
This approach eliminates the need for humans to operate the GUI directly while ensuring the entire interaction paradigm remains interpretable.

To examine whether existing OS agents can identify untrustworthy scenarios, proactively query humans, and follow query-answer pairs for trustworthy task completion, we first conduct a pilot study. 
For capability evaluation, we further construct VeriOS-Bench, a cross-platform benchmark covering mobile, desktop, web, and tablet environments. It includes annotations on untrustworthy scenario types and contains query-answer pairs that incorporate the agents' query history along with corresponding human responses.
The pilot study shows that the judgment accuracy of popular OS agents for scenario types ranged only from 46.52\% to 60.96\%. Moreover, we find that directly incorporating the query-answer history between the agent and humans into the prompt resulted in a decline in the agents’ performance on normal scenarios.

To enable OS agents to accurately judge the current scenario type, ask queries appropriately in untrustworthy scenarios, and improve their performance in such scenarios without compromising their ability in normal scenarios, we propose a three-stage learning paradigm that facilitates the decoupling and utilization of meta-knowledge by supervised fine-tuning (SFT) and group relative policy optimization (GRPO).
In this framework, each data instance in VeriOS-Bench is decoupled into meta-knowledge components: (i) scenario knowledge for identifying untrustworthy scenarios and generating corresponding queries, and (ii) action knowledge for generating executable actions using query-answer pairs. 

In the SFT stage, each type of knowledge is subjected to interleaved training, simultaneously learning the logical connections between posing queries and the generation of actions using query-answer pairs, resulting in VeriOS-Agent-SFT.

In the GPRO stage, we designed a rule-based reward system for VeriOS-Agent-SFT. This includes a scenario reward to help VeriOS-Agent more accurately determine when to ask questions, and an action reward to ensure the agent asks questions precisely and adheres to the interaction history.

Therefore, VeriOS-Agent is capable of both interacting with humans by asking queries in untrustworthy scenarios and leveraging interaction history to output GUI actions for executing user tasks. 
In untrustworthy scenarios, VeriOS-Agent asks the user for clarification before proceeding, while in normal scenarios, it automatically executes the task.

Experimental results show that VeriOS-Agent achieves an average improvement of 20.64\% in step-wise success rate on untrustworthy scenarios compared to the strongest baseline, without degrading normal scenario performance. 

Additional analysis and discussion also validate the rationality of VeriOS-Agent’s architectural design and training strategy. 
% We find that the single agent obtained through our two-stage learning paradigm performs comparably to a dual-agent system (i.e., comprising a scenario judgment agent and an action agent) constructed with the same parameter scale.
Moreover, we observe that for OS agents, interleaved training with samples from different types of knowledge yields the best learning performance.
Out-of-distribution (OOD) experiments and scalability studies further demonstrate that VeriOS-Agent exhibits strong generalization and scalability.

In summary, we make four key contributions:

(i) We propose a human-agent-GUI interaction method where the OS agent queries humans and leverages answer histories to execute tasks in untrustworthy scenarios, while autonomously performing tasks in normal scenarios.  

(ii) We propose a three-stage learning paradigm that facilitates the decoupling and utilization of meta-knowledge by SFT and GPRO, seamlessly integrating different types of knowledge into a single OS agent during training.

(iii) We introduce VeriOS-Bench, a cross-platform benchmark covering mobile, desktop, web, and tablet environments with annotations on untrustworthy scenario types and query-answer pairs. 

(iv) We contribute VeriOS-Agent, trustworthy OS agent capable of accurately assessing scenarios and accomplishing tasks through human queries. Experiments show that VeriOS-Agent achieves an average improvement of 19.72\% in step-wise success rate compared to the strongest baseline.
VeriOS-Agent significantly improves performance in untrustworthy scenarios while maintaining comparable performance in trustworthy scenarios.

\section{Related Work}
In this section, we first introduce the development of OS Agents based on MLLMs. 
% We then discuss the issue of untrustworthy components in such agents, before finally reviewing related work on systems incorporating human-in-the-loop designs.
Then, we introduce how existing work incorporates Human-in-the-loop in OS Agents.
\subsection{MLLM-based OS Agents}
As MLLMs continue to advance, MLLM-based OS agents can automatically execute user instructions by operating a GUI on smart devices such as smartphones~\cite{wu2025quick, li2025mobileuse}, computers~\cite{xie2024osworld, bonattiwindows}, and web~\cite{he2024webvoyager,zheng2024gpt} applications.
These MLLM-based OS agents can be categorized into two types based on their construction methods: single agents~\cite{qin2025ui, wuatlas, gu2025ui, zhang2025agentcpm} and multi-agent systems~\cite{ye2025mobile, zhang2025appagent}. 
For single agents, existing work primarily enhances the OS agent's capabilities in grounding~\cite{tang2025gui, li2025screenspot, chen2025v2p, zhou2025gui, liu2025infigui} and reasoning through pre-training~\cite{wu2025smoothing,  hong2024cogagent}, SFT~\cite{ma2024coco, zhang2024you, meng2024vga}, reinforcement learning (RL)~\cite{liu2025infiguir1, lu2025ui,luo2025gui, bai2024digirl, wang2024distrl, lai2025computerrl} and test-time scaling~\cite{yang2025gta1, wu2025dimo, shen2025thinking}. 
For multi-agent systems, current efforts involve constructing planning~\cite{wang2024mobile, xu2025aguvis}, reflection~\cite{li2025mobileuse, zheng2024gpt}, and memory agents~\cite{zhang2025appagent, wang2024agent,agashe2025agent} to form a multi-agent system that leverages the strengths of different agents to complete tasks. 
% These tasks are already capable of being automated to a certain extent, effectively executing user commands in an ideal environment.
However, most of these works focus on the automated execution of tasks in trustworthy scenarios, overlooking the potential risks of over-execution in untrustworthy scenarios.

% \subsection{Untrustworthy components in OS agents}

% For OS agents, there is a lot of work focused on addressing specific untrustworthy components in the environment.
% Some studies~\cite{lu2025eva, yan2025lasm} have respectively used attention distribution detection and hierarchical parameter methods to reveal how to defend against indirect prompt injection attacks and pop-ups in the environment.
% Other studies~\cite{ye2025visualtrap, cheng2025hidden} have revealed potential hidden risks of backdoor attacks in the execution environment of OS agents and attempted to use direct preference optimization methods for defense.
% Meanwhile, since human instructions may not always be absolutely precise, some studies~\cite{wu2025quick, wang2025perpilot, zhao2025appagent} have utilized users' historical operation trajectories or information databases to filter out untrustworthy components caused by vague instructions.
% However, these efforts often propose targeted methods for a specific type of untrustworthy component. 
% To universally mitigate the risks posed by untrustworthy scenarios, the OS agent must align with human intentions. 
% When the OS agent has access to historical human-related information, it can infer potential human intentions from this data~\cite{wu2025quick,jones2025users}.
% However, when the OS agent lacks additional human information, human-in-the-loop involvement becomes necessary to achieve intention alignment through clarification.

\subsection{Human-in-the-loop for OS Agents}
The execution environment for OS agents is not always an ideal environment; hence, some tasks employ a human-in-the-loop approach to enhance the agents' adaptability to non-ideal environments.
CowPilot~\cite{huq-etal-2025-cowpilot} employs manual judgment to determine whether human intervention is needed in human-agent interaction. 
To automatically identify the timing for human intervention, OS-Kairos~\cite{cheng2025kairos} allows agents to output both actions and their confidence scores, requesting human intervention when the score falls below a threshold.
GEM~\cite{wu2025gem} utilizes out-of-distribution (OOD) detection to request human interaction when a sample is identified as an OOD sample. 
% However, when OS-Kairos and GEM request human interaction, humans may not understand why the interruption occurs.
To enhance the explainability of human-agent interaction, Navi-plus~\cite{cheng2025navi} lets agents query humans in response to ambiguous user instructions, while InquireMobile~\cite{ai2025inquiremobile} employs a training approach to teach agents to engage in interactive questioning with humans during exceptional scenarios on mobile platforms.
However, there is a lack of research aimed at systematically studying the paradigm of human-in-the-loop OS agents in untrustworthy scenarios.

\begin{table*}[htb]
\centering
\caption{Pilot study on the step-wise success rate of the OS agent for normal scenarios after adding query-answer pairs.}
\label{tab:qa_pair_study}
\begin{tabular}{p{1.9cm}p{3cm}p{3cm}p{3cm}p{3cm}}
\toprule
\textbf{Method} & \textbf{Qwen2.5-VL-3B} & \textbf{Qwen2.5-VL-7B} & \textbf{Qwen2.5-VL-32B} & \textbf{Qwen2.5-VL-72B} \\
\midrule
Zero-shot & 3.66 & 37.80 & 60.98 & 71.95 \\
+ QA pair & 2.44 & 30.49 & 60.98 & 70.73 \\
\hline
\textbf{Change} & \textbf{-33.33\%} & \textbf{-19.34\%} & \textbf{0.00\%} & \textbf{-1.70\%} \\
\bottomrule
\end{tabular}
\end{table*}

\section{Investigating the Challenge of Human-Agent-GUI interaction for OS Agents}
In this section, we first introduce how we construct VeriOS-Bench, then present the problem formulation for OS agents, and finally illustrate the challenge of human-agent-GUI interaction for OS agents through a pilot study.

\subsection{VeriOS-Bench}
First, we provide a detailed introduction to why we built VeriOS-Bench and how we construct VeriOS-Bench.

Although the ability of OS agents to automatically execute user instructions on smart terminal devices is continuously improving, current OS agents focus on the automated execution of commands in an ideal environment. 
Yet their operating conditions are often not necessarily trustworthy~\cite{yang2025mla, yu2025survey, shi2025towards}.
As shown in figure~\ref{fig:scenario}, these trust issues stem from two types of factors: environment-side and user-side. 
Environment-side factors include environmental anomalies caused by OS instability~\cite{ma2024caution} (e.g., disconnections, OS crashes) and inherently sensitive actions~\cite{cheng2025kairos} within the environment (e.g., triggering certain controls that may release location permissions). 
User-side factors involve information missing due to ambiguous user instructions~\cite{cheng2025navi,xu2025mobile} and situations with multiple choices~\cite{peng2025morae}.

To better evaluate the performance of existing OS agents in untrustworthy scenarios, we first collect 622 screenshots of untrustworthy scenarios and normal scenarios across mobile, desktop, web, and tablet through crowdsourcing and existing OS agent benchmarks~\cite{cheng2025kairos,wu2025quick,ma2024caution,wang2025mmbench,zhang2024android,kapoor2024omniact,cheng2024seeclick,wuatlas,li2025screenspot,lu2024gui}. 
To comprehensively evaluate the OS agents, we collect screenshots of untrustworthy and normal scenarios in a ratio of approximately 6 to 4.

Then, we employ expert annotation to label the instructions, ground truth, and scenario types for these screenshots. 
The action space of VeriOS-Bench encompasses common GUI operations such as click, swipe, long press, and type. 
Additionally, we use a crowdsourcing method to annotate, for each screenshot belonging to untrustworthy scenarios, the content that the OS agent should ask the human to complete the task, along with the predefined human stance.

\begin{figure}[t]
    \centering        
    \includegraphics[width=0.48\textwidth]{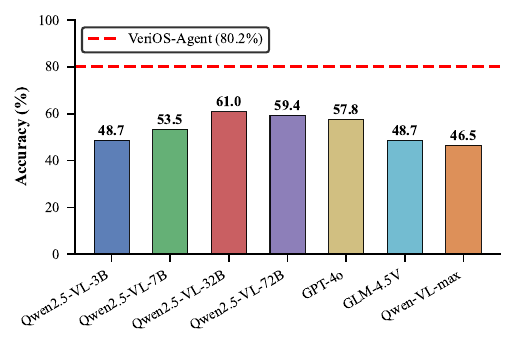} 
    \caption{Pilot study on the scenario judgment accuracy of normal MLLM-based OS agents. Existing MLLM-based OS agents perform poorly in identifying untrustworthy scenarios.}
    \label{fig:pilot} 
\end{figure}

\subsection{Pilot Study of existing OS Agents}

We conduct a pilot study on normal MLLM-based OS agents on VeriOS-Bench. 
Concretely, we first test existing common MLLM-based OS agents to perform five-category classification of scenarios (environmental anomalies, information missing, sensitive actions, multiple choices, and normal scenarios), and we report the scenario judgment accuracy, which is defined as the proportion of correct scenario identifications by the agent.
Results in Figure~\ref{fig:pilot} show that the judgment accuracy of existing OS agents ranges only from 46.52\% to 60.96\%, indicating that existing OS agents fail to accurately identify untrustworthy scenarios. In contrast, VeriOS-Agent (which we will introduce later) can improve scenario judgment accuracy to 80.21\%.

To test whether existing OS agents can leverage query-answer history to better accomplish tasks, we directly incorporate correctly annotated query-answer pairs from VeriOS-Bench into the prompt.
As shown in Table~\ref{tab:qa_pair_study}, we find that after adding the query-answer pairs, the step-wise success rate of the OS agent in normal scenarios decreased to varying degrees, with a more pronounced decline observed in smaller-scale models. 
This indicates that OS agents in a zero-shot setting struggle to utilize human-agent interaction query-answer pairs to accomplish tasks effectively and may even impair their original task completion capabilities.

Achieving trustworthy Human-Agent-GUI interaction remains a non-trivial challenge. 
Based on the analysis above, we identify two primary factors contributing to this difficulty for OS agents: (i) current OS agents are unable to accurately detect untrustworthy scenarios, and (ii) even when provided with entirely correct query–answer histories, their performance degrades in otherwise normal task settings.
These observations highlight the need for an OS agent capable of accurately recognizing the current task scenario type, generating appropriate queries to the user, and effectively leveraging user responses to improve task execution. 
This motivates the design of VeriOS-Agent, which we introduce in the next section.

\begin{figure*}[t]
    \centering        
    \includegraphics[width=0.95\textwidth]{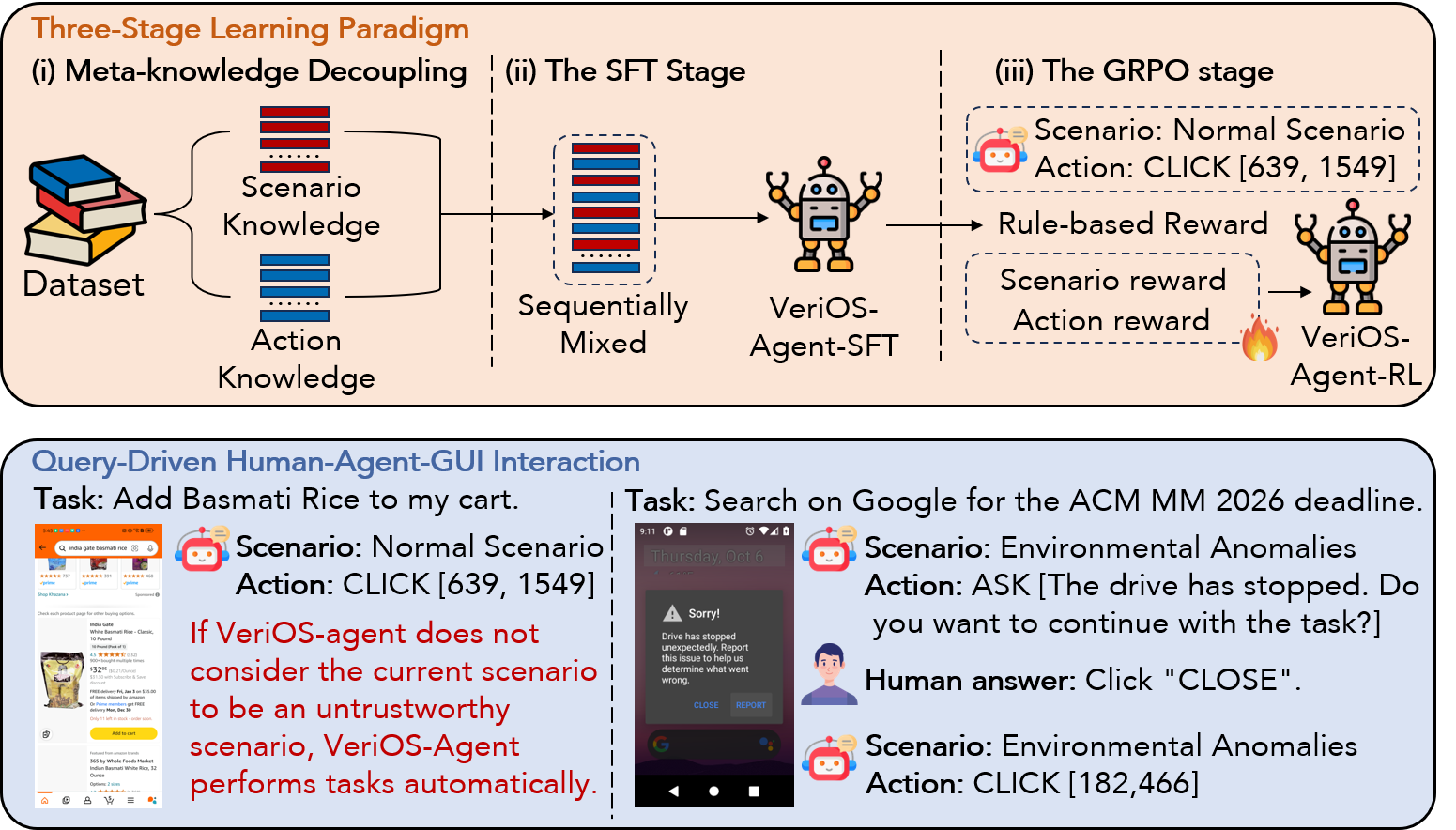} 
    \caption{Diagram of the two-stage learning paradigm and query-driven human-agent-GUI interaction. The two-stage learning paradigm consists of the meta-knowledge decoupling stage and the meta-knowledge utilization stage. We first decouple the knowledge from VeriOS-Bench into scenario knowledge and action knowledge, and then leverage this knowledge to construct VeriOS-Agent. During the interaction process, when VeriOS-Agent identifies the current scenario as untrustworthy, it issues a query to the human and utilizes the history of queries and human responses to better accomplish the task.}
    \label{fig:method} 
\end{figure*}

\section{VeriOS-Agent}
In this section, we first introduce how we train VeriOS-Agent using a three-stage learning paradigm that facilitates the decoupling and utilization of meta-knowledge by SFT and GRPO, and then we explain how to use VeriOS-Agent for the proactive query-driven human-agent-GUI interaction. 
Figure~\ref{fig:method} shows the overall framework.

\subsection{Framework Overview}

\subsubsection{Meta-knowledge decoupling stage}
In meta-knowledge decoupling stage, each data instance \(d\) in VeriOS-Bench \(\mathcal{D}\) is defined as a tuple:
\(d = (P, i, s, a_g, e, q, h),\)
where \(P\) denotes the system prompt, \(i \in \mathcal{I}\) denotes the user instruction, \(s\) denotes the screenshots, \(a_g \in \mathcal{A}\) denotes the ground truth action, \(e\) denotes the scenario type, \(q\) denotes the standard ask query, and \(h\) denotes the standard human answer. Note that \(q = h = \emptyset\) when \(e\) corresponds to a normal scenario.

Each instance encapsulates two distinct forms of knowledge: (i) scenario knowledge for identifying untrustworthy scenarios and generating corresponding asks, denoted as \(\mathcal{K}_{\text{scenario}}\), and (ii) action knowledge for generating executable actions using query-answer pairs, denoted as \(\mathcal{K}_{\text{action}}\). 
To enable the OS agent to learn both types of knowledge effectively, we propose a three-stage learning paradigm that facilitates the decoupling and utilization of meta-knowledge.

For any instance \(d \in \mathcal{D}\), we decouple it into two derived sub-instances: \(d_1\) and \(d_2\). The first sub-instance, \(d_1\), is designed to elicit \(\mathcal{K}_{\text{scenario}}\), and is constructed as:
\begin{equation}
d_1: \text{Input} = \langle P, i, s\rangle, \quad \text{Output} = \langle e, q \rangle.
\end{equation}

The second sub-instance, \(d_2\), is designed to elicit \(\mathcal{K}_{\text{action}}\), and is constructed as:
\begin{equation}
d_2: \text{Input} = \langle P, i, s, q, h \rangle, \quad \text{Output} = \langle e, a_g \rangle.
\end{equation}

The training dataset \(\mathcal{D}_{\text{train}}\) is then constructed by interleaving the sub-instances from each original instance in \(\mathcal{D}\). Specifically, for each \(d \in \mathcal{D}\), we include \(d_1\) followed by \(d_2\) in \(\mathcal{D}_{\text{train}}\), ensuring that the two types of knowledge are learned in an interleaved manner. Formally, if the original dataset is ordered as \(d^{(1)}, d^{(2)}, \ldots, d^{(N)}\), then:
\begin{equation}
\mathcal{D}_{\text{train}} = \left( d_1^{(1)}, d_2^{(1)}, d_1^{(2)}, d_2^{(2)}, \ldots, d_1^{(N)}, d_2^{(N)} \right).
\end{equation}

\begin{table}[t]
\centering
\small
\setlength{\tabcolsep}{3.2pt}         
\caption{
  Experimental results of different baselines on VeriOS-Bench.
  MC: multiple choices; IM: information missing; EA: environment anomalies;
  SA: sensitive actions; NS: normal scenarios; Total: overall step-wise success rate.}
\label{tab:model_performance}
\begin{tabular}{lcccccc} 
\toprule
\textbf{Model} & \textbf{MC} & \textbf{IM} & \textbf{EA} & \textbf{SA} &
\textbf{NS} & \textbf{Total} \\ 
\midrule
UI-TARS-2B-SFT & 16.67 & 14.29 & 45.45 & 66.67 & 41.46 & 37.97 \\
Qwen2.5-VL-3B & 0.00 & 0.00 & 22.73 & 7.41 & 3.66 & 5.35 \\
Qwen2.5-VL-7B & 2.38 & 0.00 & 0.00 & 44.44 & 37.80 & 23.53 \\
OS-Atlas-Pro-7B & 19.05 & 7.14 & 22.73 & 59.26 & 64.63 & 44.39 \\
UI-TARS-7B-SFT & 21.43 & 21.43 & 59.09 & 70.37 & 59.76 & 49.73 \\
UI-TARS-7B-DPO & 19.05 & 14.29 & 54.55 & 70.37 & 63.41 & 49.73 \\
UI-TARS-1.5-7B & 23.81 & 14.29 & 59.09 & 62.96 & 58.54 & 48.13 \\
Qwen2.5-VL-32B & 9.52 & 14.29 & 36.36 & 77.78 & 60.98 & 45.45 \\
GUI-Owl-32B & 14.29 & 14.29 & 54.55 & 70.37 & 69.51 & 51.34 \\
Qwen2.5-VL-72B & 16.67 & 21.43 & 50.00 & 77.78 & \textbf{71.95} & 54.01 \\
GPT-4o & 14.29 & 7.14 & 31.82 & 81.48 & 48.78 & 40.64 \\
Qwen-VL-max & 11.90 & 14.29 & 4.55 & 18.52 & 26.83 & 18.72 \\
GLM-4.5V & 23.81 & 21.43 & 50.00 & 77.78 & 63.41 & 51.87 \\
\midrule
\rowcolor{gray!20}
VeriOS-Agent-7B-SFT
& \cellcolor{gray!20}45.24 & \cellcolor{gray!20}71.43 & \cellcolor{gray!20}77.27
& \cellcolor{gray!20}81.48 & \cellcolor{gray!20}47.56 & \cellcolor{gray!20}57.22 \\

\rowcolor{gray!40}
VeriOS-Agent-32B-SFT
& \cellcolor{gray!40}45.24 & \cellcolor{gray!40}71.43 & \cellcolor{gray!40}81.82
& \cellcolor{gray!40}92.95 & \cellcolor{gray!40}69.51 & \cellcolor{gray!40}68.98 \\

\rowcolor{gray!20}
VeriOS-Agent-7B-RL
& \cellcolor{gray!20}46.43 & \cellcolor{gray!20}71.43 & \cellcolor{gray!20}79.55
& \cellcolor{gray!20}81.48 & \cellcolor{gray!20}48.17 & \cellcolor{gray!20}65.41 \\

\rowcolor{gray!40}
VeriOS-Agent-32B-RL
& \cellcolor{gray!40}47.62 & \cellcolor{gray!40}73.21 & \cellcolor{gray!40}84.09
& \cellcolor{gray!40}94.12 & \cellcolor{gray!40}70.12 & \cellcolor{gray!40}73.83 \\

\bottomrule
\end{tabular}
\end{table}
\subsubsection{SFT stage}
In the SFT stage, interleaved SFT is performed over \(\mathcal{D}_{\text{train}}\). The training objective for a sub-instance \(d_1\) with target token sequence \(y^{(1)} = \langle e, q \rangle\) is:
\begin{equation}
\mathcal{L}_{d_1} = -\sum_{t=1}^{L_1} \log p\left(y_t^{(1)} \mid P, i, y_{<t}^{(1)}; \theta\right),
\end{equation}
where \(L_1\) is the length of \(y^{(1)}\) and \(\theta\) represents the model parameters. Similarly, for a sub-instance \(d_2\) with target sequence \(y^{(2)} = \langle e, a_g \rangle\), the loss is:
\begin{equation}
\mathcal{L}_{d_2} = -\sum_{t=1}^{L_2} \log p\left(y_t^{(2)} \mid P, i, q, h, y_{<t}^{(2)}; \theta\right),
\end{equation}
where \(L_2\) is the length of \(y^{(2)}\).

The overall training objective is the average loss over all sub-instances:
\begin{equation}
\mathcal{L}_{\text{SFT}}(\theta) = \frac{1}{|\mathcal{D}_1|} \sum_{d_1 \in \mathcal{D}_1}\mathcal{L}_{d_1} + \frac{1}{|\mathcal{D}_2|} \sum_{d_2 \in \mathcal{D}_2}\mathcal{L}_{d_2}.
\end{equation}

\subsubsection{GRPO stage}
In the GRPO stage, we further refine the policy of \textit{VeriOS-Agent} using Group Relative Policy Optimization (GRPO).
For each input instance $d$, the agent samples a group of $G$ candidate outputs $\{o_1, o_2, \ldots, o_G\}$, which are evaluated using a multi-dimensional reward function.

\paragraph{Reward Function Design.}
The overall reward $R$ is designed to jointly capture scenario understanding and action execution quality, consisting of two components:

\textbf{Scenario Reward ($r_{\text{sce}}$).}
This term evaluates the correctness of scenario perception and query generation.
Formally, $r_{\text{sce}} = 1$ if both the predicted scenario $\hat{e}$ matches the ground truth $e$ and the generated query $\hat{q}$ matches $q$; otherwise, $r_{\text{sce}} = 0$.

\textbf{Action Reward ($r_{\text{act}}$).}
This term measures the quality of the executable action:
\begin{equation}
r_{\text{act}} = \frac{r_{\text{type}} + r_{\text{acc}}}{2},
\end{equation}
where $r_{\text{type}} = 1$ if the predicted action type is correct (and $0$ otherwise), and $r_{\text{acc}} = 1$ if all action parameters are exactly correct (and $0$ otherwise).

The final reward is computed as:
\begin{equation}
R = \frac{r_{\text{sce}} + r_{\text{act}}}{2}.
\end{equation}

\paragraph{Optimization Objective.}
GRPO optimizes the policy by maximizing the relative advantage of samples within each group:
\begin{equation}
\begin{aligned}
\mathcal{J}_{\text{GRPO}}(\theta)
= \mathbb{E} \Bigg[
\frac{1}{G} \sum_{i=1}^{G}
\Big(
\min \Big(
\frac{\pi_{\theta}(o_i|d)}{\pi_{\text{old}}(o_i|d)} \hat{A}_i,
\text{clip}(\cdot)
\Big) \\
- \beta \, \mathbb{D}_{\text{KL}}(\pi_{\theta} \,\|\, \pi_{\text{ref}})
\Big)
\Bigg]
\end{aligned}
\end{equation}
where the normalized advantage $\hat{A}_i$ is computed as:
\begin{equation}
\hat{A}_i =
\frac{R_i - \mathrm{mean}(R_1, \ldots, R_G)}
{\mathrm{std}(R_1, \ldots, R_G)}.
\end{equation}

\subsection{Query-Driven Human-Agent-GUI Interaction}

When a human provides an instruction \(i\) to the VeriOS-Agent \(M\) at time step \(t\), \(M\) first captures a screenshot \(s_t\) of the current interface. VeriOS-Agent then determines the type of the current scenario and generates an action:
\begin{equation}
e_t, a_{t}^{(0)} = M(i, s_t),
\end{equation}
where \(e_t\) denotes the judgment of the scenario type at time \(t\), and \(a_{t}^{(0)}\) represents the output action.

If the scenario type \(e_t\) is normal, \(a_{t}^{(0)}\) must not be an ASK action but rather an action that can directly operate the GUI. In this case, the final executed action is \(a_t = a_{t}^{(0)}\).  
If the scenario type \(e_t\) is one of untrustworthy scenarios, \(a_{t}^{(0)}\) must be the ASK action, prompting the VeriOS-Agent to pose a clarifying query to the human.

Subsequently, the human provides a response \(h_t\) to the query raised by \(a_{t}^{(0)}\).  
VeriOS-Agent then performs another round of reasoning using both the query \(a_{t}^{(0)}\) and the human response \(h_t\):
\begin{equation}
e_t, a_{t}^{(1)} = M(i, s_t, a_{t}^{(0)}, h_t).
\end{equation}

To prevent the agent from repeatedly asking queries, \(a_{t}^{(1)}\) must not be an ASK action when \(h_t\) is non-empty.  
The final action executed at time step \(t\) is then \(a_t = a_{t}^{(1)}\).  
After applying \(a_t\) to the environment, the state transitions to \(s_{t+1}\). The task terminates if either the objective is completed or the maximum number of steps \(T\) set by the human is reached. Otherwise, the process repeats for time step \(t+1\).

\begin{table*}[ht]
\centering
\setlength{\tabcolsep}{3.2pt}
\caption{%
  Comparison of single-agent vs.\ dual-agent systems with 7B scale \textit{(left)} and 32B scale \textit{(right)}.
  The dual-agent system yields only marginal gains in overall (\textbf{Total}) and lower scenario-judgment accuracy (\textbf{SJA}).}
\label{tab:single_or_dual}
\begin{tabular}{lccccccclccccccc}
\toprule
& \multicolumn{7}{c}{\textbf{7B scale}} && \multicolumn{7}{c}{\textbf{32B scale}} \\
\cmidrule{2-8}\cmidrule{10-16}
\textbf{Method} & MC & IM & EA & SA & NS & Total & SJA && MC & IM & EA & SA & NS & Total & SJA \\
\midrule
\rowcolor{gray!20}
Single-agent
& \cellcolor{gray!20}45.24 & \cellcolor{gray!20}\textbf{71.43} & \cellcolor{gray!20}\textbf{77.27}
& \cellcolor{gray!20}\textbf{81.48} & \cellcolor{gray!20}47.56 & \cellcolor{gray!20}57.22
& \cellcolor{gray!20}\textbf{80.21}
&& \cellcolor{gray!20}45.24 & \cellcolor{gray!20}\textbf{71.43} & \cellcolor{gray!20}\textbf{81.82}
& \cellcolor{gray!20}\textbf{92.95} & \cellcolor{gray!20}\textbf{69.51} & \cellcolor{gray!20}\textbf{68.98}
& \cellcolor{gray!20}\textbf{74.87} \\
Dual-agent system   & \textbf{50.00} & 42.86 & 72.73 & 74.07 & \textbf{56.10} & \textbf{58.29} & 77.54
             && \textbf{50.00} & \textbf{71.43} & 72.73 & 92.59 & 62.20 & 65.78 & 70.05 \\
\bottomrule
\end{tabular}
\end{table*}

\begin{table*}[t]
\centering
\setlength{\tabcolsep}{3.2pt}         
\caption{%
  Comparison of training methods with Qwen2.5-VL-7B \textit{(left)} and 32B \textit{(right)}.
  Interleaved Training consistently yields the best.}
\label{tab:learn_method}
\begin{tabular}{lccccccclccccccc}
\toprule
& \multicolumn{7}{c}{\textbf{Qwen2.5-VL-7B}} && \multicolumn{7}{c}{\textbf{Qwen2.5-VL-32B}} \\
\cmidrule{2-8}\cmidrule{10-16}
\textbf{Method} & MC & IM & EA & SA & NS & Total & SJA && MC & IM & EA & SA & NS & Total & SJA \\
\midrule
Shuffled Learning        & 30.95 & \textbf{71.43} & 54.55 & 55.56 & 32.93 & 41.18 & \textbf{80.75}
               && 42.86 & \textbf{71.43} & \textbf{81.82} & \textbf{92.59} & 64.63 & 66.31 & 74.87 \\
\rowcolor{gray!20}
Interleaved Learning
& \cellcolor{gray!20}\textbf{45.24} & \cellcolor{gray!20}\textbf{71.43} & \cellcolor{gray!20}\textbf{77.28}
& \cellcolor{gray!20}\textbf{81.48} & \cellcolor{gray!20}\textbf{47.56} & \cellcolor{gray!20}\textbf{57.22}
& \cellcolor{gray!20}80.21
&& \cellcolor{gray!20}\textbf{45.24} & \cellcolor{gray!20}\textbf{71.43} & \cellcolor{gray!20}\textbf{81.82}
& \cellcolor{gray!20}\textbf{92.59} & \cellcolor{gray!20}\textbf{69.51} & \cellcolor{gray!20}\textbf{68.98}
& \cellcolor{gray!20}74.87 \\
Rotating Learning       & 11.90 & 0.00 & 22.73 & 51.85 & 29.27 & 25.67 & 80.21
               && \textbf{47.62} & 64.29 & \textbf{81.82} & 88.89 & 65.85 & 66.84 & \textbf{78.61} \\
Phased Learning         & 19.05 & 7.14 & 45.45 & 44.44 & 31.71 & 30.48 & 47.06
               && 28.57 & 28.57 & 59.09 & 81.48 & 64.63 & 55.61 & 59.36 \\
\bottomrule
\end{tabular}
\end{table*}

\section{Experiments}
In this section, we first introduce the experimental setup, followed by the main results and findings on VeriOS-Bench. 
\subsection{Experiments Setup}
\subsubsection{Baselines}
We select a total of 13 popular OS agents as baselines, including those built on open-source models and those based on proprietary models. 
The open-source model-based agents include the UI-TARS~\cite{qin2025ui} series, Qwen2.5-VL~\cite{Qwen2.5-VL} series, OS-Atlas-Pro-7B~\cite{wuatlas}, UI-TARS-1.5-7B, and GUI-Owl-32B~\cite{ye2025mobile}.
Among these, all except Qwen2.5-VL are specialized models in the field of OS agents.
The proprietary model-based agents include GPT-4o~\cite{hurst2024gpt}, Qwen-VL-max~\cite{Qwen-VL}, and GLM-4.5V~\cite{vteam2025glm45vglm41vthinkingversatilemultimodal}.
% Since GPT-4o inherently lacks the capability to predict coordinates, we incorporate an OCR model composed of ResNet18~\cite{he2016deep} and ConvNeXt-Tiny~\cite{liu2022convnet} to assist them in localization.
The base OS agent models fine-tuned under our framework are based on Qwen2.5-VL-7B and Qwen2.5-VL-32B.

\subsubsection{Metrics}
In the experiments of this paper, we evaluate whether the output actions of the OS agent match the ground truth and report the step-wise success rate under different scenarios, as well as the overall step-wise success rate. 
Here, MC refers to multiple choices, IM refers to information missing, EA refers to environment anomalies, SA refers to sensitive actions, and NS refers to normal scenarios. 
We also report the scenario judgment accuracy (SJA) to measure the accuracy of the OS agent's scenario judgment. 
For the step-wise success rate of each individual scenario and the overall total, we report the proportion of actions generated by the OS agent that are correct.
For SJA, we report the proportion of scenario types correctly identified by the OS agent.

% \subsubsection{Implementation Details}
% During the training process, we performed full fine-tuning on both the visual and LLM components. The VeriOS-Agent-7B model was trained with a learning rate of 5e-6 for two epochs, while the VeriOS-Agent-32B model was trained with a learning rate of 1e-6 for two epochs. Throughout training, we set the maximum image resolution to 262144 pixels and used a batch size of 1 per GPU. 
% Following mainstream evaluation practices, when calculating evaluation metrics, a relative error of up to 14\% is allowed for click and long press actions compared to the ground truth. For type actions, the input text must achieve a similarity of at least 80\% with the ground truth text. All other actions must exactly match the ground truth to be considered correct.
% The base OS agent models fine-tuned under this framework are based on Qwen2.5-VL-7B and Qwen2.5-VL-32B. The resulting agents are denoted as VeriOS-Agent-7B and VeriOS-Agent-32B.

\subsection{Main Results}

To verify the trustworthiness of existing OS agents and to assess whether VeriOS-Agent is trustworthy, we conducted experiments on VeriOS-Bench.

As shown in Table~\ref{tab:model_performance}, we have the following key findings:  
(i) Existing popular OS agents are not trustworthy. 
When facing untrustworthy scenarios, the step-wise success rate of the baselines' actions is very low, particularly in cases involving user-induced multiple choices and information missing, where the step-wise success rate ranges only from 0\% to 23.81\%.  
(ii) VeriOS-Agent is somewhat trustworthy. Both VeriOS-Agent-7B and VeriOS-Agent-32B significantly improve the ability to handle untrustworthy scenarios.  
(iii) While enhancing the capability to address untrustworthy scenarios, VeriOS-Agent maintains its performance in handling normal scenarios without any decline. 
This indicates that the diversity of training data helps improve the generalization of OS agents.
\begin{figure*}[t]
    \centering        
    \includegraphics[width=1\textwidth]{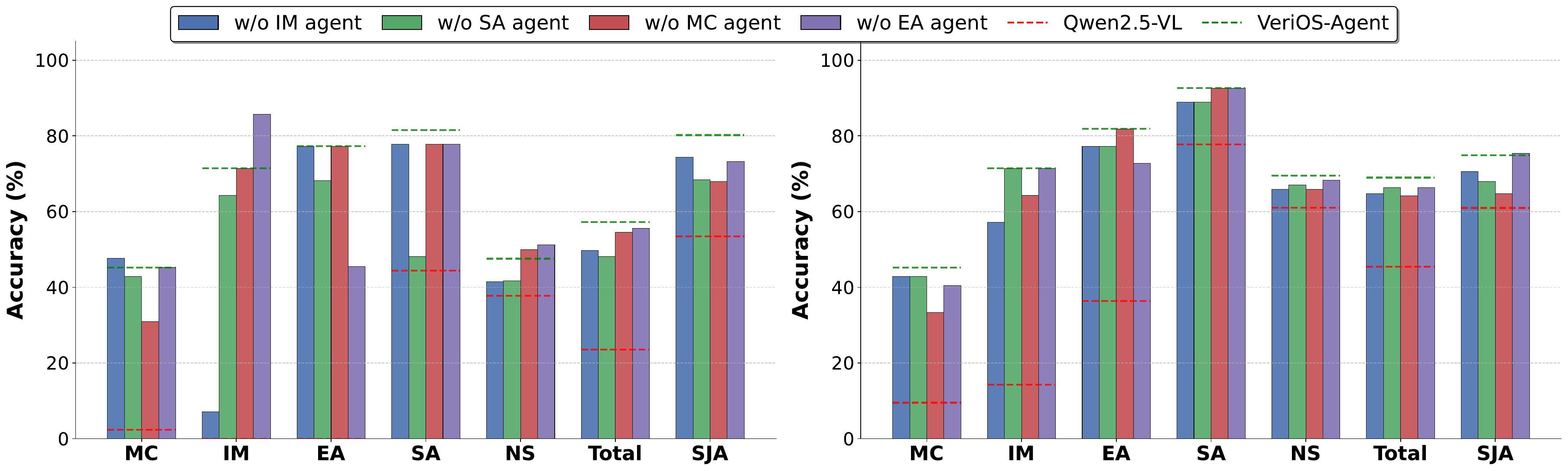} 
    \caption{OOD experiment with 7B and 72B model parameter scales. Experimental results demonstrate that VeriOS-Agent exhibits strong generalization capabilities.}
    \label{fig:ood} 
\end{figure*}

\begin{table}[t]
\centering
\setlength{\tabcolsep}{6pt}
\caption{Scalability study of VeriOS-Agent. Experimental results demonstrate the good scalability of VeriOS-Agent.}
\label{tab:scalability_study}

\begin{tabular}{p{3cm}ccc}
\toprule
\textbf{Model} & \textbf{NS} & \textbf{Total} & \textbf{SJA} \\
\midrule
\rowcolor{gray!20}
VeriOS-Agent-32B-SFT
& \cellcolor{gray!20}\textbf{69.51} & \cellcolor{gray!20}68.98 & \cellcolor{gray!20}74.87 \\
SA, MC, EA→IM & 68.29 & \textbf{70.05} & 72.19 \\
IM, MC, EA→SA & 65.85 & 67.38 & 76.47 \\
SA, IM, EA→MC & 64.63 & 68.45 & 66.84 \\
SA, MC, IM→EA & 68.29 & 67.91 & \textbf{77.01} \\
\bottomrule
\end{tabular}
\end{table}

\begin{table*}[ht]
\centering
\setlength{\tabcolsep}{3.2pt}
\caption{%
  Comparison of experimental results using different base models for training (\textbf{left}: 7B scale, \textbf{right}: 32B scale).  
  Experiments show that the general-purpose Qwen2.5-VL series performs better as the base model.}
\label{tab:model_comparison}
\begin{tabular}{lccccccclccccccc}
\toprule
& \multicolumn{7}{c}{\textbf{7B scale}} && \multicolumn{7}{c}{\textbf{32B scale}} \\
\cmidrule{2-8}\cmidrule{10-16}
\textbf{Model} & MC & IM & EA & SA & NS & Total & SJA && MC & IM & EA & SA & NS & Total & SJA \\
\midrule
\rowcolor{gray!20}
Qwen2.5-VL
& \cellcolor{gray!20}\textbf{45.24} & \cellcolor{gray!20}\textbf{71.43} & \cellcolor{gray!20}\textbf{77.27}
& \cellcolor{gray!20}\textbf{81.48} & \cellcolor{gray!20}\textbf{47.56} & \cellcolor{gray!20}\textbf{57.22}
& \cellcolor{gray!20}\textbf{80.21}
&& \cellcolor{gray!20}\textbf{45.24} & \cellcolor{gray!20}\textbf{71.43} & \cellcolor{gray!20}\textbf{81.82}
& \cellcolor{gray!20}\textbf{92.95} & \cellcolor{gray!20}\textbf{69.51} & \cellcolor{gray!20}\textbf{68.98}
& \cellcolor{gray!20}\textbf{74.87} \\
GUI-Owl     & 23.81 & 50.00 & 77.27 & 74.07 & 31.71 & 42.78 & 73.26
            && 28.57 & 64.29 & 77.27 & 92.95 & 65.85 & 62.57 & 73.80 \\
\bottomrule
\end{tabular}
\end{table*}

\section{Analysis and Discussion}
In this section, we provide a detailed analysis of the rationale behind the construction of VeriOS-Agent and demonstrate its generalization capability and scalability through OOD experiments and a scalability study.
Unless otherwise specified, the subject of analysis throughout this section is VeriOS-Agent-SFT.
\subsection{Why not Adopt a Dual-Agent System?}

We employ a three-stage learning paradigm to decouple human-agent-GUI interaction capabilities at the knowledge level. 
Intuitively, we can also perform agent-level decoupling by separately fine-tuning two OS agents—one agent specializes in determining scenario types and posing queries, while the other focuses on generating actions.

As shown in Table~\ref{tab:single_or_dual}, we observe that the dual-agent system does not yield significant improvements compared to the single agent. 
This is because there exist inherent logical connections among scenario type determination, query posing, and action generation. 
And the single agent can implicitly learn these connections.
Moreover, the dual-agent system requires greater memory overhead and more computational resources for both training and inference.
So agent-level decoupling is unnecessary for our task setting.

\subsection{How to Arrange Decoupled Knowledge?}

We decouple each data point $d$ into $d_1$ and $d_2$. 
However, during SFT for the OS agent, we have multiple ways of arranging $d_1$ and $d_2$. 
Ultimately, we adopt the interleaved training approach (i.e., first training on the two decoupled data points from one sample in the training set, then moving to the next sample). 
However, we also consider the following methods:
(i) Shuffled training: Training on completely randomized data within the same epoch.
(ii) Rotating training: For each epoch, training on all data of the first meta-knowledge first, followed by all data of the other meta-knowledge.
(iii) Phased training: First, train sufficiently on all data of the first meta-knowledge for multiple epochs, then train on all data of the other meta-knowledge.

As shown in Table~\ref{tab:learn_method}, we find that interleaved training yields the best experimental results. 
This is because interleaved training takes into account the inherent logical relationship between the two types of meta-knowledge.
while shuffled training completely discards this information. 
In contrast, both rotating training and phased training lead to catastrophic forgetting of different meta-knowledge in the OS agent.
We also find that the performance of the 32B scale agent is less affected by the training method compared to the 7B scale agent. 
This is because a larger number of parameters represents more world knowledge and stronger reasoning capabilities, making it easier to learn the implicit logical relationships between different training data without relying on the explicit learning of logical relationships through the order of data training.

\subsection{Base Model Selection Experiment}

In the main results, we observe that GUI-Owl outperforms Qwen2.5-VL at the same parameter scale. 
However, the Qwen2.5-VL series is selected as base model. It is therefore essential to validate whether this choice is justified.

To justify our base model selection, we conduct a base model selection experiment. As shown in Table~\ref{tab:model_comparison}, GUI-Owl's foundational capabilities are not as strong as those of Qwen2.5-VL. 
This is because GUI-Owl is a specialized model for the OS agent domain, while Qwen2.5-VL is a generalized model. 
The judgment of scenario types and responses to human queries in OS agent tasks requires a certain level of general language understanding, making Qwen2.5-VL a more effective choice as the base model.

\subsection{OOD Experiment}
To test the generalization of VeriOS-Agent, we conduct an OOD experiment. 
We sequentially remove the data related to each untrustworthy scenario from the training set to examine whether VeriOS-Agent exhibits certain generalization abilities on OOD scenarios. 
The experimental results are shown in Figure~\ref{fig:ood}. 
We observe that even though knowledge of some untrustworthy scenarios is not learned during the OOD experiment, the performance of each model on every metric surpasses the red line. 
This indicates that VeriOS-Agent possesses a certain degree of generalization ability after learning the judgment and "ask" knowledge from other untrustworthy scenarios. 
This is highly valuable for the real-world applicability of VeriOS-Agent.

\subsection{Scalability Study}

To explore the scalability of this work, we assume that VeriOS-Agent has been trained on data from three types of untrustworthy scenarios, but a new set of data containing a fourth type of untrustworthy scenario emerges, introducing new knowledge. 
This setup simulates the scalability challenges that VeriOS-Agent might encounter in the real world. 
Therefore, we sequentially exclude the knowledge of each untrustworthy scenario from the original training set to train the model, followed by a two-stage training process using the excluded data. 
As shown in Table~\ref{tab:scalability_study}, the agent trained in two stages still achieves the same performance comparable to VeriOS-Agent, and even surpasses VeriOS-Agent in some instances. 

This demonstrates the exceptional scalability of this work. 
Essentially, there exists an implicit logical connection among the scenario knowledge after meta-knowledge decoupling. 
The knowledge of specific scenarios already encompasses the fundamental logic of human-agent-GUI interaction, enabling VeriOS-Agent to easily extend to new scenarios and, to some extent, mitigate the catastrophic forgetting~\cite{french1999catastrophic} problem in the field of continual learning.

\section{Conclusion}
In this paper, we analyze the potential risks of over-execution by automated OS agents in untrustworthy scenarios.
To address this, we propose a query-driven human-agent-GUI interaction framework and a three-stage learning paradigm that facilitates the decoupling meta-knowledge, resulting in VeriOS-Agent.
VeriOS-Agent can query humans in untrustworthy scenarios, and automatically execute instructions otherwise.
We also introduce VeriOS-Bench, a cross-platform benchmark with untrustworthy scenario annotations. 
Experiments show that VeriOS-Agent improves step-wise success rates by 20.64\% in untrustworthy scenarios without degrading normal scenario performance.
VeriOS-Agent significantly improves performance in untrustworthy scenarios while maintaining comparable performance in trustworthy scenarios.
\bibliographystyle{ACM-Reference-Format}
\bibliography{acmart}

\appendix

\end{document}